\documentclass[sigconf]{acmart}

\usepackage{booktabs} 
\usepackage{comment}
\usepackage{graphics}
\usepackage[english]{babel}
 
\usepackage{amsthm}
 \usepackage[linesnumbered,ruled,vlined]{algorithm2e}
\usepackage{algorithmic}

\usepackage[T1]{fontenc}
\usepackage{amsmath}
\usepackage{amssymb}
\usepackage{fullpage,url}

\theoremstyle{definition}
\newtheorem{definition}{Definition}[section]

\newtheorem{lemma}{Lemma}
\newtheorem{example}{Example}

\acmDOI{00}

\acmConference[]{}{}{}
\copyrightyear{2019}

\begin{document}
\title{Efficient Matrix Profile Computation Using Different Distance Functions}

\author{Reza Akbarinia}
\affiliation{%
  \institution{INRIA \& LIRMM,  Univ. Montpellier, France}
}
\email{reza.akbarinia@inria.fr}

\author{Bertrand Cloez}
\affiliation{%
  \institution{INRA, SupAgro, UMR MISTEA, Univ. Montpellier, Montpellier, France}
 }
\email{bertrand.cloez@inra.fr}

\renewcommand{\shortauthors}{R. Akbarinia et al.}

\begin{abstract}

Matrix profile has been recently proposed as a promising technique to the problem of all-pairs-similarity search on time series. Efficient algorithms have been proposed for computing it, \textit{e.g.}, STAMP \cite{YZUB+2016}, STOMP \cite{ZZSY+2016} and SCRIMP++ \cite{ZhuYZKK18}. All these algorithms use the z-normalized Euclidean distance to measure the distance between subsequences. However, as we observed, for some datasets other Euclidean measurements are more useful for knowledge discovery from time series. 

In this paper, we propose efficient algorithms for computing matrix profile for a general class of Euclidean distances. We first propose a simple but efficient algorithm called AAMP for computing matrix profile with the "pure" (non-normalized) Euclidean distance. Then, we extend our algorithm for the p-norm distance. We also propose an algorithm, called ACAMP, that uses the same principle as AAMP, but for the case of z-normalized Euclidean distance. 
We implemented our algorithms, and evaluated their performance through experimentation. The experiments show excellent performance results. For example, they show that AAMP is very efficient for computing matrix profile for non-normalized Euclidean distances. The results also show that the ACAMP algorithm is significantly faster than SCRIMP++ (the state of the art matrix profile algorithm) for the case of z-normalized Euclidean distance.

\end{abstract}

%
%

\maketitle

\section{Introduction}

Matrix profile has been recently proposed as en efficient technique to the problem of all-pairs-similarity search in time series \cite{YHK2016, LZPK2018, DK2017, YKK2017, ZINK2017}. Given a time series $T$ and a subsequence length $m$, the matrix profile returns for each subsequence included in $T$ its distance to the most similar subsequence in the time series. The matrix profile is itself a time series very useful for data analysis, \textit{e.g.}, detecting the motifs (represented by low values), discords (represented by high values), etc. 

One naive solution for computing a matrix profile is the nested loop algorithm that for each subsequence computes its distance to any other subsequence using the distance function. However, this solution is not efficient and can take too much time for relatively big databases. Recently, efficient algorithms have been proposed for matrix profile computation, \textit{e.g.}, STAMP \cite{YZUB+2016}, STOMP \cite{ZZSY+2016} and SCRIMP++ \cite{ZhuYZKK18}. 
All these algorithms use the z-normalized Euclidean distance to measure the distance between subsequences. 

\begin{figure}
\centering

\includegraphics[scale=0.23]{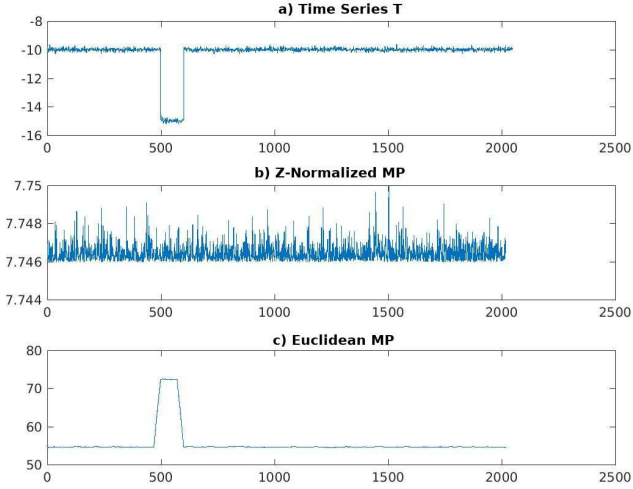}

  \caption{\label{example-euclidean} a) Example of a time series $T$; b) matrix profile of $T$ generated using z-normalized Euclidean distance; c) matrix profile of $T$ generated using pure Euclidean distance}
\end{figure}

However, we observed that for some datasets, other distances such as \textit{pure} (non-normalized) Euclidean distance are more useful for knowledge discovery. For example, in the case of time series containing long subsequences of the same values (that show some type of stability in the activities), the z-normalized distance cannot be used because the standard deviation of these subsequences is zero, thus the z-normalized distance becomes infinite. In addition, in some cases the normalization can remove some rare information from the matrix profile. As an example, consider Figure \ref{example-euclidean}.a that shows a time series $T$, and Figures  \ref{example-euclidean}.b and  \ref{example-euclidean}.c that depict the matrix profiles generated from $T$ using z-normalized and pure Euclidean distances respectively. As seen, the matrix profile that uses z-normalized distance looses the information about the anomaly around 500 in the time series. But, the pure Euclidean distance highlights it.

We believe that for knowledge extraction from different datasets, we need to give to the users the possibility of computing matrix profiles with different similarity distances. In this paper, we propose matrix profile algorithms for different types of Euclidean distance. Our contributions are as follows. 
\begin{itemize}

\item We first propose a simple but efficient algorithm called AAMP for computing matrix profile with the pure Euclidean distance. AAMP is executed in a set of iterations, such that in each iteration the distance of subsequences is computed incrementally. The time complexity of AAMP is $O(n \times (n - m))$ with small constants, where $n$ is the time series length and $m$ the subsequence length.  

\item We extend AAMP to compute matrix profile for the p-norm distance that is more general than Euclidean.

\item We propose an extension of AAMP, called ACAMP, that uses the same principle as AAMP but for z-normalized Euclidean distance. In ACAMP, we use an incremental formula
for computing z-normalized distance that is based on some variables computed incrementally in a sliding window that moves over the subsequences of the time series.
\end{itemize}

We precise that these new algorithms are exact, anytime and incrementally maintainable. They take a deterministic execution time that only depends on the time series and subsequence length.

We implemented our algorithms and compared them with the state of the art algorithm on matrix profile, \textit{i.e.}, SCRIMP++ \cite{ZhuYZKK18}. The results show excellent performance gains. For example, they show that the execution time of AAMP for pure Euclidean and p-norm distances is several times smaller than that of SCRIMP++. Also, they show that the ACAMP algorithm can outperform SCRIMP++ with a factor of more than 50\%. 

The rest of this paper is organized as follows. In Section \ref{sec_problem_definition}, we give the problem definition. In Section  \ref{sec_aamp}, we describe our AAMP algorithm for computing matrix profile with pure Euclidean and p-norm distances. In Section \ref{sec_acamp}, we propose the ACAMP algorithm for z-normalized distance.
Section \ref{sec_experiments} presents the experimental results. Section \ref{sec_related_work} discusses related work, and 
Section \ref{sec_conclusion} concludes.

\section{Problem Definition}\label{sec_problem_definition} 
In this section, we give the formal definition of the matrix profile, and describe the problem we address. 

\theoremstyle{definition}
\begin{definition}
A \textit{time series} $T$ is a sequence of real-valued numbers $T = \langle t_1, \dots, t_n \rangle$ where $n$ is the length of $T$. 
\end{definition}

A subsequence of a time series is defined as follows. 
\theoremstyle{definition}
\begin{definition}
 Let $m$ be a given integer value such that $ 1 \leq m \leq n$. A \textit{subsequence} $T_{i, m}$  of a time series $T$ is a continuous sequence of  values in $T$ of length $m$ starting from position $i$. Formally, $T_{i, m} = \langle t_i, \dots, t_{i+m-1}\rangle $ where $ 1 \leq i \leq n - m + 1$. We call $i$ the start position of $T_{i, m}$.
\end{definition}

For each subsequence of a time series we can compute its distance to all
subsequences of the same length in the same time series. We call this a distance profile.

\theoremstyle{definition}
\begin{definition}
\label{def:dist}
 Given a query subsequence $T_{i, m}$, a \textit{distance profile} $D_i$ of $T_{i, m}$ in the time series $T$ is a
vector of the distances between $T_{i, m}$  and each subsequence of length $m$ in time series $T$.
Formally, $D_i = \langle d_{i,1}, \dots, d_{i,n-m+1} \rangle$, where $d_{i,j}$ is the distance between $T_{i,m}$ and $T_{j,m}$.
\end{definition}

Note that the term distance in Definition~\ref{def:dist} does not refer to the mathematical definition of a distance. It only gives a measure on the difference between two subsequences. For instance the z-normalized Euclidean distance does not satisfy the (mathematical) axioms of a distance.


A \textit{matrix profile} is a vector that represents the minimum distance of each subsequence of $T$ to other subsequences of $T$.

\theoremstyle{definition}
\begin{definition}
Given a length $m$, the \textit{matrix profile} of a time series $T$ is a vector $P = \langle p_1, \dots, p_{n-m+1} \rangle$ such that $p_i$ is the minimum distance of the subsequence $T_{i,m}$ to any other subsequence of $T$, for $1<i<n-m+1$. In other words, $p_i = min (D_i)$, \textit{i.e.}, $p_i$ is the minimum value in the distance profile of $T_{i,m}$.
\end{definition}

In this paper, we are interested in efficient computation of matrix profile using three different distance measures: 1) Euclidean distance; 2) p-norm distance that is a generalization of Euclidean distance; 3) z-normalized Euclidean distance. These distances are defined as follows. 

\theoremstyle{definition}
\begin{definition}
The \textit{Euclidean distance} between two subsequences $T_{i,m}$ and $T_{j,m}$ is defined as:
\begin{equation} \label{def-euclidean-distance}
D_{i,j} = \sqrt{\sum_{l=0}^{m-1} (t_{i+l} - t_{j+l})^2}
\end{equation}
\end{definition}

In this paper, sometimes we call the Euclidean distance as \textit{pure Euclidean distance}.

\theoremstyle{definition}
\begin{definition}
Let $p > 1$ be a positive integer, then the \textit{p-norm distance} between two subsequences $T_{i,m}$ and $T_{j,m}$ is defined as:
\begin{equation} \label{def-p-norm-distance}
DP_{i,j} = \sqrt[p]{\sum_{l=0}^{m-1} (t_{i+l} - t_{j+l})^p}
\end{equation}
\end{definition}

The z-normalized Euclidean distance is defined as follows.

\theoremstyle{definition}
\begin{definition}
Let $\mu_i$ and $\mu_j$ be the mean of the values in two subsequences $T_{i,m}$ and $T_{j,m}$ respectively. Also, let $\sigma_i$ and $\sigma_j$ be the standard deviation of the values in $T_{i,m}$ and $T_{j,m}$ respectively. Then, the \textit{z-normalized Euclidean distance} between $T_{i,m}$ and  $T_{j,m}$ is defined as:

\begin{equation} \label{def-z-normalized-distance}
DZ_{i,j}= \sqrt{\sum_{l=0}^{m-1} \left(\frac{t_{i+l} - \mu_i}{\sigma_i}- \frac{t_{j+l} - \mu_j}{\sigma_j} \right)^2}
\end{equation}

\end{definition}


\begin{figure*}[t]
\centering
\includegraphics[scale=0.35]{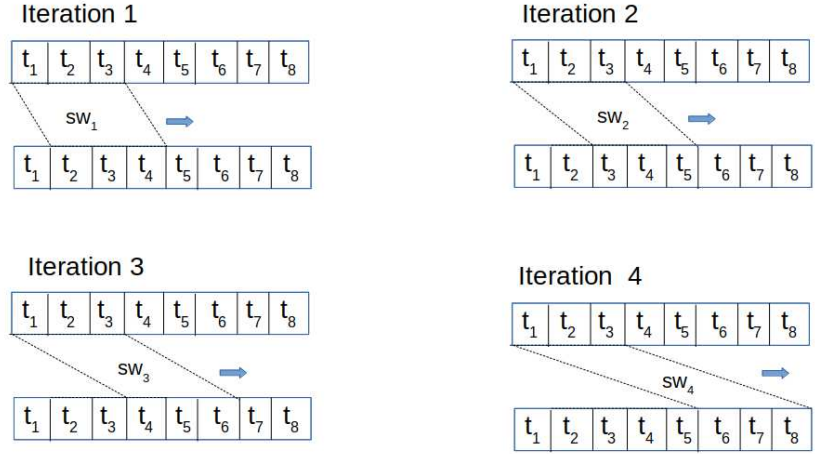}
\caption{Example of AAMP execution on a time series of length n =8, and with subsequence length m=3. In each iteration $k$, in a sliding window subsequences are compared with those that are $k$ positions far from them}
\label{fig-aamp-example}
\end{figure*}

\section{AAMP}\label{sec_aamp}
In this section, we propose AAMP an efficient algorithm for computing matrix profile using the Euclidean distance. We first present a formula for incremental computation of the Euclidean distance in $O(1)$, and then we detail our AAMP algorithm that uses this formula for computing matrix profile. 

\subsection{Incremental Computation of Euclidean Distance}

Here, we present a formula that allows us to compute the Euclidean distance between two subsequences $T_{i,m}$ and $T_{j,m}$ based on the Euclidean distance of subsequences $T_{i-1,m}$ and $T_{j-1,m}$.  The formula is presented by the following lemma.

\begin{lemma}\label{lemma-inc-euclidean}
Let $D_{i,j}$ be the Euclidean distance between two subsequences $T_{i,m}$ and $T_{j,m}$. Let  $D_{i-1,j-1}$ be the Euclidean distance between two subsequences $T_{i-1,m}$ and $T_{j-1,m}$. Then $D_{i,j}$ can be computed as:

\begin{equation} \label{IncrementalEucDist}
D_{i,j} = \sqrt{D_{i-1,j-1}^2 - (t_{i-1} - t_{j-1})^2 + (t_{i+m-1} - t_{j+m-1})^2}
\end{equation}

\end{lemma}

\textbf{Proof.} Let $T_{i, m} = \langle t_i, t_{i+1}, \dots, t_{i+m-1}\rangle$ and $T_{j, m} = \langle t_j, t_{j+1},$  $ \dots, t_{j+m-1}\rangle$. Then the square of the Euclidean distance between  $T_{i,m}$ and $T_{j,m}$ is computed as:

\begin{equation} \label{EucDist1}
D_{i,j}^2 = \sum_{l=0}^{m-1} (t_{i+l} - t_{j+l})^2
\end{equation}

And the square of the Euclidean distance between $T_{i-1,m}$ and $T_{j-1,m}$ is:

\begin{equation} \label{EucDist2}
D_{i-1,j-1}^2 = \sum_{l=0}^{m-1} (t_{i-1+l} - t_{j-1+l})^2
\end{equation}

By comparing Equations \eqref{EucDist1} and \eqref{EucDist2}, we have:

$D_{i,j}^2 = D_{i-1,j-1}^2 - (t_{i-1} - t_{j-1})^2 + (t_{i+m-1} - t_{j+m-1})^2 $. Thus, we have:

$D_{i,j} = \sqrt{D_{i-1,j-1}^2 - (t_{i-1} - t_{j-1})^2 + (t_{i+m-1} - t_{j+m-1})^2}$.

$\Box$

By using the above equation, we can compute the Euclidean distance $D_{i,j}$ by using the distance $D_{i-1,j-1}$ in $O(1)$. 

\subsection{Algorithm}


The main idea behind AAMP is that for computing the distance between subsequences it uses \textit{diagonal sliding windows}, such that in each sliding window, the Euclidean distance is computed only between the subsequences that have a precise difference in their \textit{start position}. These sliding windows allow us to use Equation \eqref{IncrementalEucDist} for efficient distance computation. 



Algorithm \ref{aamp_algo} shows the pseudo-code of AAMP. 
Initially, the algorithm sets all values of the matrix profile to infinity (\textit{i.e.}, maximum distance). Then, it performs $n-m - 1$ iterations using a variable $k$ ($ 1 \leq k \leq n - m - 1$). In each iteration $k$, the algorithm compares each subsequence $T_{i,m}$ with the subsequence that is $k$ positions far from it, \textit{i.e.}, $T_{i,m+k}$. To do this, AAMP firstly computes the Euclidean distance of the first subsequence of the time series, \textit{i.e.}, $T_{1,m}$, with the one that stars at position $k$, \textit{i.e.}, $T_{k,m}$. This first distance computation is done using the normal formula of Euclidean distance, \textit{i.e.}, that of Equation \eqref{def-euclidean-distance}. Then, in a sliding window, the algorithm incrementally computes the distance of other subsequences with the subsequences that are $k$ position far from them, and this is done by using Equation \eqref{IncrementalEucDist} in $O(1)$. If the computed distance is smaller than the previous minimum distance that is kept in the matrix profile $P$, then the smaller distance is saved in the matrix profile. 

\begin{example}
Figure \ref{fig-aamp-example} shows an example of executing AAMP over a time series of length $n=8$, and for subsequences of length $m=3$. In this example, the algorithm proceeds in $4$ iterations ($n-m-1 = 4$). In Iteration 1, firstly the Euclidean distance between $T_{1, m}$ and $T_{2, m}$ is calculated using the normal Euclidean distance formula. Then the sliding window $sw_1$ moves to the next subsequences, and computes the distance of $T_{2, m}$ and $T_{3, m}$ using Equation \eqref{IncrementalEucDist}. Then, the sliding window moves to the next subsequences and computes their distances, \textit{i.e.}, $T_{3, m}$ and $T_{4, m}$. This continues until computing the distance of subsequences that have one point of difference in their start position. In the second iteration, in the sliding window $sw_2$, the Euclidean distance is computed between each subsequence and the one that is "two" positions far from it. This continues until Iteration 4. Note that in each iteration the first distance is computed using the normal formula of Euclidean distance, and the other distances are computed using the incremental formula, \textit{i.e.}, Equation \eqref{IncrementalEucDist}.
\end{example}

As an optimization of AAMP, we can use the square of the Euclidean distance for comparing the distance of different subsequences, and at the end of the algorithm replace the square of the distance by the real distance in the matrix profile. This optimization reduces the number of sqrt operations done during the algorithm execution.

\begin{algorithm}
\DontPrintSemicolon 
\KwIn{$T$: time series; $n$:  length of time series; $m$: subsequence length}
\KwOut{$P$: Matrix profile; }

\Begin{

\For{i=1 to n}{
    P[i] = $\infty$ ; // initialize the matrix profile
    } 

\For{k=1 to n-m-1}{
    $dist = Euc\_Distance (T_{1,m}, T_{k, m}$) // compute the distance between $T_{1,m}$, $T_{k, m}$ 
   
   \If{dist < P[1]}{
        P[1] = dist;
    }
    
    \If{dist < P[k]}{
        P[k] = dist;
        
   }
    
    \For{i=2 to n - m + 1 - k}{
    
        $dist = \sqrt{(dist^2 -  (t_{i-1} - t_{i - 1 + k})^2 + (t_{i+m-1} - t_{i + m + k - 1} )^2 }$ 
        
        \If{dist < P[i]}{
            P[i] = dist;
            
        }
        
        \If{dist < P[i+k]}{
            P[i+k] = dist;
            
        }
    } 
  }
}

\caption{AAMP algorithm: matrix profile with Euclidean distance \label{aamp_algo}}	
\end{algorithm}

\subsection{Complexity Analysis} \label{sec-complexity-AAMP}
Here, we analyze the time and space complexity of AAMP. The algorithm contains two loops. In the first loop, in Line 5 the distance between $T_{1,m}$ and $T_{k,m}$ is computed using the normal Euclidean distance function in $O(m)$, thus in total Line 5 is executed in $O(m \times (n - m))$. In the nested loop (Lines 10-15), all operations are done in $O(1)$, so in total these operations are done in $O((n - m)^2)$. Thus, the time complexity of the algorithm is $O((n - m)^2) + m \times (n - m))$ that is equivalent of $O(n \times (n - m))$. If $m$ is small compared to $n$, \textit{i.e.}, $n>>m$, then the time complexity of AAMP can be written as $O(n^2)$. But, if $m$ is very close to $n$, \textit{i.e.},  $m = n - c $ for a small constant $c$, then the time complexity is $O(n)$.


The space needed for executing our algorithm is only the array of matrix profile and some simple variables. Thus, the space complexity of AAMP is $O(n)$.

\subsection{Extension of AAMP to p-Norm Distance}
In this section, we extend the AAMP algorithm to the p-norm distance that is a more general distance than Euclidean. The p-norm functions are used in Lebesgue spaces ($L^P$), which are useful in data analysis in physics, statistics, finance, engineering, etc.

Let $T_{i,m}$ and $T_{j,m}$ be two time series subsequences, then their p-norm distance (for $ p \geq 1 $) is defined as:

\begin{equation} \label{PNormDist1}
DP_{i,j} = \sqrt[p]{\sum_{l=0}^{m-1} (t_{i+l} - t_{j+l})^p}
\end{equation}

Notice that the Euclidean distance is a special case of p-norm with $p=2$. 

The following lemma gives an incremental formula for computing $PNORM_{i,j}$.

\begin{lemma}\label{lemma-inc-p-norm}
Let $DP_{i,j}$ be the p-norm distance of subsequences $T_{i,m}$ and $T_{j,m}$. Then, $DP_{i,j}$ can be computed by using the p-norm distance of subsequences $T_{i-1,m}$ and $T_{j-1,m}$, denoted by $DP_{i-1,j-1}$, as following:

$DP_{i,j} = $ 

$\sqrt[p]{(DP_{i-1,j-1})^p - (t_{i-1} - t_{j-1})^p + (t_{i+m-1} - t_{j+m-1})^p}$

\end{lemma}

\textbf{Proof.} The proof can be easily done in a similar way as that of Lemma \ref{lemma-inc-euclidean}.  
$\Box$

Using Lemma \ref{lemma-inc-p-norm}, we can modify the AAMP algorithm to compute the matrix profile with the p-norm distance. This can be done just by modifying two lines in Algorithm \ref{aamp_algo}: 1) Line 5 : by replacing the Euclidean distance with the p-norm distance of subsequences  $T_{1,m}$ and $T_{k,m}$; 2) Line 11: incrementally computing the p-norm distance using the equation of Lemma \ref{lemma-inc-p-norm}.  

The pseudo-code of AAMP algorithm for the p-norm distance is shown in Appendix. The time and space complexity of the AAMP algorithm for p-norm is the same as that of AAMP with the Euclidean distance. 

\section{ACAMP: Matrix Profile for Z-Normalized Euclidean Distance}\label{sec_acamp}
In this section, we propose an algorithm, called ACAMP, that computes matrix profile based on the z-normalized euclidean distance and using the same principle as AAMP, \textit{i.e.}, incremental distance computation in diagonal sliding windows. However, the incremental computation of the distance in ACAMP is different than that of AAMP.

\subsection{Incremental Computation of Z-Normalized Euclidean Distance}
Let us now explain how ACAMP computes the z-normalized Euclidean distance incrementally. Let $T_{i, m} = \langle t_{i}, \dots, t_{i+m-1}\rangle$ and  $ T_{j,m} =\langle t_{j}, \dots, t_{j+m-1} \rangle$ be two subsequences of a time series $T$. In ACAMP, we compute the z-normalized Euclidean distance between $T_{i, m}$ and $T_{j, m}$ using the  following five variables: 

\begin{itemize}
\item $A_i = \sum_{l=0}^{m-1} t_{i+l}$: the sum of the values in $ T_{i,m}$; 

\item $B_j = \sum_{l=0}^{m-1} t_{j+l}$: the sum of the values in  $ T_{j,m}$; 

\item  $\mathbf{A_i} = \sum_{l=0}^{m-1} t_{i+l}^2$: the sum of the square of values in $T_{i,m}$; 

\item  $\mathbf{B_j} = \sum_{l=0}^{m-1} t_{j+l}^2$: the sum of the square of values in $T_{j,m}$; 

\item  $\mathbf{C_{i,j}} = \sum_{l=0}^{m-1} t_{i+l} \times t_{j+l}$: the product of values of $T_{i,m}$ and  $T_{j,m}$. 


\end{itemize}

Note that all above variables can be computed incrementally, when moving a sliding window from  $T_{i,m}$ to  $ T_{i+1,m}$. Given these variables, then the z-normalized Euclidean distance between two subsequences $T_{i,m}$ and  $T_{j,m}$ can be computed using the formula given by the following lemma.

\begin{lemma}\label{lemma_inc-z-normalized}
Let $DZ_{i,j}$ be the z-normalized distance of subsequences $T_{i,m}$ and $T_{j,m}$. Then, $DZ_{i,j}$ can be computed as:

\begin{equation} \label{IncrementalZnormal}
DZ_{i,j} = \sqrt{2m \left(1 - \frac{\mathbf{C_{i,j}} - \frac{1}{m}A_i B_j }{\sqrt{\left(\mathbf{A_i} - \frac{1}{m}A_{i}^2\right)\left(\mathbf{B_j} - \frac{1}{m}B_{j}^2\right)}}\right)}
\end{equation}

\end{lemma}

\textbf{Proof.} The proof can be seen in Appendix.

\subsection{Algorithm} \label{sect:acamp_algo}
The pseudo-code of ACAMP is shown in Algorithm \ref{acamp_algo}. After initializing the matrix profile, ACAMP performs $n-m-1$ iterations, such that in iteration $k$ it compares the z-normalized Euclidean distance of subsequences that are $k$ points far from each other in the time series (Lines 4 to 13). In each iteration, the distances are computed using the formula of Equation  \ref{IncrementalZnormal} using the five variables which are used in the equation, \textit{i.e.}, $A_i$, $B_j$, $\mathbf{A}_i$, $\mathbf{B}_j$ and $\mathbf{C_{i,j}}$. For the first subsequences of the iteration, \textit{i.e.}, $T_{1,m}$ and $T_{1+k,m}$, the variables are computed using their normal formula in $O(m)$ (see Lines 5 to 9). For the other subsequences of the iterations, these variables are computed incrementally, \textit{i.e.}, in $O(1)$.

Note that in the algorithm, for performance reasons we compare the square of the z-normalized Euclidean distance of the subsequences (Line 10 and 21). By this, we avoid performing $O(n^2)$ sqrt operations in our nested loop. At the end of the algorithm (Lines 26 to 27), in a loop we convert the square distances to the real distances, using $O(n)$ sqrt operations.

\begin{algorithm}
\DontPrintSemicolon
\KwIn{T: time series; n:  length of time series; m: subsequence length}
\KwOut{P: Matrix profile; }

\Begin{

\For{i=1 to n}{
    P[i] = $\infty$ ; // initialize the matrix profile
    } 

\For{k=1 to n-m+1}{
    $A = \sum_{l=0}^{m-1} t_{1+l}$ //sum of the values in $T_{1,m}$; 

    $B = \sum_{l=0}^{m-1} t_{1+k+l}$: // sum of the values in  $ T_{1+k,m}$; 

    $\mathbf{A} = \sum_{l=0}^{m-1} t_{1+l}^2$: // sum of the square of values in $T_{1,m}$; 

    $\mathbf{B} = \sum_{l=0}^{m-1} t_{1+k+l}^2$: // sum of the square of values in $T_{1+k,m}$; 

    $\mathbf{C} = \sum_{l=0}^{m-1} t_{1+l} t_{k+l}$: // product of values of $T_{1,m}$ and  $T_{1+k,m}$.
    
    $dist =2m \left(1 - \frac{\mathbf{C} - \frac{1}{m}A B }{\sqrt{(\mathbf{A} - \frac{1}{m}A^2)(\mathbf{B} - \frac{1}{m}B^2)}} \right)$
    // compute the square of z-normalized distance
   
   \If{dist < P[1]}{
        P[1] = dist;
    }
    
    \If{dist < P[k]}{
        P[k] = dist;
        
   }
    
    \For{i=2 to n - m + 1 - k}{
        
        $A = A - t_{i-1} +  t_{i+m-1}$;
        
         $B = B - t_{i-1 + k} +  t_{i + m + k - 1}$;
        
         $\mathbf{A} =  \mathbf{A} - t_{i-1}^2 +  t_{i+m-1} ^2$;
         
         $\mathbf{B} = \mathbf{B} - t_{i-1 + k}^2 +  t_{i + m + k - 1}^2$;
         
         $\mathbf{C} = \mathbf{C} - t_{i-1} \times t_{i-1 + k} + t_{i+m-1} \times t_{i + m + k - 1}$;
        
        $dist =2m \left(1 - \frac{\mathbf{C} - \frac{1}{m}A B }{\sqrt{(\mathbf{A} - \frac{1}{m}A^2)(\mathbf{B} - \frac{1}{m}B^2)}} \right)$
        
        \If{dist < P[i]}{
            P[i] = dist;
            
        }
        
        \If{dist < P[i+k]}{
            P[i+k] = dist;
            
        }
    } 
  }
  
  \For{i=1 to n}{
    $P[i] = \sqrt{P[i]}$ ; // compute the z-normalize distance from its square
    } 
  
}

\caption{ACAMP algorithm: matrix profile calculation with z-normalized Euclidean distance \label{acamp_algo}}	
\end{algorithm}

\subsection{Complexity Analysis} \label{sec-complexity-ACAMP}
Let us now analyze the time and space complexity of ACAMP. The algorithm proceeds in two loops. In the first loop the variables needed for computing the distance (Lines 5 to 9) are computed in $O(m)$, thus in total this part of the algorithm is executed in $O(m \times (n-m))$. In the nested loop, the variables are computed in $O(1)$, thus in total the Lines 16 to 25 are done in $O((n-m)^2)$. Therefore, the time complexity of the algorithm is $O(n \times (n-m))$. If $m$ is small compared to $n$, then the time complexity of ACAMP is $O(n^2)$. But, if $m$ is very close to $n$, \textit{i.e.},  $m = n - c $ for a small constant $c$, then the time complexity of ACAMP is linear, \textit{i.e.}, $O(n)$.


The algorithm needs to keep only some variables and an array for the output matrix profile, thus its space complexity is $O(n)$, \textit{i.e.}, the size of the output.

\subsection{More Optimization of ACAMP} \label{acamp_optimizations}

We can further optimize ACAMP by not comparing the square of z-normalized distance in Lines 11, 13, 22 and 24 in Algorihtm~\ref{acamp_algo}, but by comparing $F_{i,j}$ defined as follows:  
\begin{equation}
\label{eq:F}
F_{i,j} = \frac{ (A_{i} B_j -m \mathbf{C_{i,j}}) \times |   A_{i} B_j -m \mathbf{C_{i,j}}| }{( \mathbf{A}_i-  \frac{1}{m} A_i^2)( \mathbf{B}_j- \frac{1}{m} B_j)},
\end{equation}

We can easily show that $DZ_{i,j} > DZ_{i,k}$ if and only if $F_{i,j}> F_{i,k}$. In the formula of $F_{i,j}$, there is no sqrt operation, and its computation takes less time than that of $DZ_{i,j}$. Thus, for comparing the z-normalized Euclidean distance of subsequences, we can simply compare their $F_{i,j}$. Then in Line 27 of the algorithm, the following equation can be used for computing the z-normalized Euclidean distance $DZ_{i,j}$ from $F_{i,j}$:  
\begin{equation}
\label{eq:F-to-DZ}
DZ_{i,j}=2m + 2 \text{sign}(F_{i,j}) \times \sqrt{|F_{i,j}|}
\end{equation}

Another possible optimization is to move the first calculation of variables $A$, $\mathbf{A}$, $B$, and $\mathbf{B}$ (actually done in Lines 5 to 8) before the loop (\textit{i.e.},  before Line 4), and incrementally update these variables in the loop.

\begin{figure}
\centering

\includegraphics[scale=0.60]{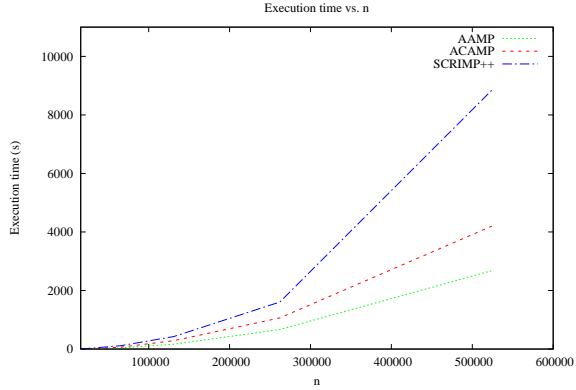}

  \caption{\label{cout-m-fixe} Execution time of the three algorithms when the time series length $n$ varies  }
\end{figure}

\begin{figure}
\centering

\includegraphics[scale=0.60]{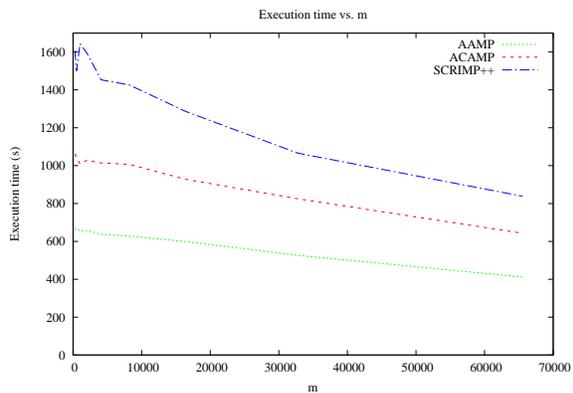}

  \caption{\label{cout-n-fixe} Execution time of the three algorithms when the subsequence length $m$ varies, and the time series length is set to $n=2^{18} \sim 262K$ }
\end{figure}

\section{Performance Evaluation}\label{sec_experiments} 

In this section, we compare the execution time of our algorithms AAMP and ACAMP with the state-of-the-art exact motif discovery algorithm SCRIMP++ \cite{ZhuYZKK18}. We first describe the experimental setup, and then present the results of our experimental evaluation.

\subsection{Setup}
We implemented our algorithms in MATLAB. For Scrimp ++, we use the algorithm available in \cite{scimppp} with step input the usual step size of PreSCRIMP which is 0.25.

The execution times of the three algorithms AAMP, ACAMP and SCRIMP++ are only affected by the length of the time series (\textit{i.e.}, $n$) and the length of the subsequences (\textit{i.e.}, $m$). The values inside the time series have no impact on the execution time of the tested algorithms, thus we generated them randomly using a uniform distribution. In our tests, we varied the parameters $n$ and $m$, and measured their impact on the algorithms execution time. Unless otherwise specified, the default values for $m$ and $n$ are $m=2^8$ and $n=2^{18}$ respectively.

The evaluation tests of the three algorithms were carried out on a machine with
Intel \textregistered Core\texttrademark i7-4770 CPU  3.40GHz ×
$\times 8$ processor, on Ubuntu 14.04 LTS and 7,7 Gio memory with the R2015B version of Matlab.  

For each test, we perform two experiments and report their average execution times.

\subsection{Results}
We studied the effect of the time series length (\textit{i.e.}, $n$) on the execution time of our algorithms.
Figure~\ref{cout-m-fixe} shows the time required by AAMP, ACAMP and SCRIMP++ to compute matrix profile for a fixed  subsequence length $m=256$, and with varying time series length values. As seen the execution time of the three algorithms increases with increasing $n$. But, AAMP and ACAMP perform much better than Scrimp++. The results show that the difference between the performance of AAMP/ACAMP and Scrimp++ increases significantly  when $n$ gets higher. 

We also studied the effect of subsequence length on the performance of our algorithms. 
Figure~\ref{cout-n-fixe} shows the execution time of the three algorithms for computing the matrix profile for time series with a fixed length of $n = 2^{18}$, and varying the subsequence length from  $m = 256$ to $ m = 2^{16}$. The response of the our algorithms and that of Scrimp++ decreases when $m$ increases. This is in accordance with our complexity analysis presented in Sections \ref{sec-complexity-AAMP} and \ref{sec-complexity-ACAMP} showing that for the cases where $m$ is close to $n$, the time complexity of our algorithms gets linear to $n$. 


\section{Related Work}\label{sec_related_work}

Motif discovery from time series is important for many application domains such as bioinformatics \cite{Sinha02}, speech processing \cite{Balasubramanian16}, Seismology \cite{YALM18} and entomology \cite{MueenKZCW09}. 
Matrix profile has been recently proposed as en efficient technique to the problem of all-pairs-similarity search on time series \cite{YHK2016, LZPK2018, DK2017, YKK2017, ZINK2017}.

In \cite{YZUB+2016}, Yeh et al. introduced the theoretical foundations of matrix profile, and proposed a first algorithm, called STAMP, for computing the matrix profile over a time series. The algorithm uses a similarity search algorithm, called MASS, that under z-normalized Euclidean computes the distance of each subsequence to other subsequences by using the Fast Fourier Transform (FFT).  
Other exact  algorithms such as Quick-Motif \cite{LiUYG15}, IMD \cite{GuHCSCH17}, or MK \cite{MueenKZCW09}  can be  fast for cooperative  data (those that are relatively smooth data, short motif lengths etc.). But in \textit{less-cooperative} data (e.g., seismology data) these algorithms are not efficient \cite{ZZSY+2016}. 

In \cite{ZZSY+2016}, Zhu et al. proposed an algorithm, called STOMP, that is faster than STAMP. The STOMP algorithm is similar to STAMP in that it can be seen as highly optimized nested loop searches, with the repeated calculation of distance profiles as the inner loop. However, while STAMP must evaluate the distance profiles in random order (to allow its anytime behavior), STOMP performs an ordered search. STOMP exploits the locality of these searches, and reduces the time complexity by a factor of $O(logn)$. 
In \cite{ZhuYZKK18}, the authors proposed an extension of STOMP, called SCRIMP++, that converges much faster than STOMP. 

To the best of our knowledge, all most all matrix profile algorithms have been developed for z-normalized Euclidean distance. In this paper, we proposed efficient algorithms for a larger class of Euclidean functions. We also proposed an algorithm for the z-normalized case, \textit{i.e.}, ACAMP, that is significantly faster than SCRIMP++, which is the fastest exact algorithm for matrix profile computation in the literature, to the best of our knowledge. Our ACAMP algorithm is designed based on an efficient incremental technique that does not need to calculate FFT (in contrast to SCRIMP++).

\section{Conclusion}\label{sec_conclusion}

In this paper, we addressed the problem of matrix profile computation for a general class of Euclidean distances. We first proposed an efficient algorithm called AAMP for computing matrix profile for the "non-normalized" Euclidean distance. Then, we extended our algorithm for the p-norm distance, which is a general form of Euclidean. Then, we proposed an algorithm, called ACAMP, that uses the same principle as AAMP, but for the case of z-normalized Euclidean distance. Our algorithms are exact, anytime, incrementally maintainable, and can be implemented easily using different languages.
To evaluate the performance of our algorithms, we implemented them, and compared their performance with the state of the art algorithm SCRIMP++. The results show excellent performance gains. For example, they show that ACAMP is significantly faster than SCRIMP++. They also show that AAMP is very efficient for computing matrix profile for non-normalized Euclidean and p-norm distances.


\bibliographystyle{ACM-Reference-Format}
\bibliography{references}

\section{Appendix A: Incremental Computation of Z-Normalized Euclidean Distance - Proof} \label{appendix-b}
Here, we present the proof of Lemma \ref{lemma_inc-z-normalized} that gives an incremental formula for computing matrix profile with z-normalized Euclidean distance.

\textbf{Proof.} Let $\mu_i$ and $\mu_j$ be the mean of the values in the sequences $T_{i,m}$ and $T_{j,m}$ respectively. Also, let $\sigma_i$ and $\sigma_j$ be the standard deviation of the values in the subsequences $T_{i,m}$ and $T_{j,m}$ respectively. Then, the z-normalized Euclidean distance between the subsequences $T_{i,m}$ and  $T_{j,m}$ is defined as:

$$
DZ_{i,j}= \sqrt{\sum_{l=1}^{m-1} \left(\frac{t_{i+l} - \mu_i}{\sigma_i}- \frac{t_{j+l} - \mu_j}{\sigma_j} \right)^2},
$$
where
$$
\mu_i= \frac{1}{m}\sum_{l=0}^{m-1} t_{i+l}, \qquad \mu_j= \frac{1}{m}\sum_{l=0}^{m-1} t_{j+l}
$$
and
$$
\sigma_i = \sqrt{ \frac{1}{m} \sum_{l=0}^{m-1} t_{i+l}^2 - (\mu_i)^2 }, \qquad \sigma_j = \sqrt{ \frac{1}{m} \sum_{k=0}^{m-1} t_{j+l}^2 - (\mu_j)^2 }.
$$

\onecolumn

We can write the square of $DZ$ as following: 	
\begin{align*}
DZ_{i,j}^2
&= \sum_{l=0}^{m-1} \left(\frac{t_{i+l} - \mu_i}{\sigma_i}- \frac{t_{j+l} - \mu_j}{\sigma_j} \right)^2\\
&= \sum_{l=0}^{m-1} \left( \left(\frac{t_{i+l} - \mu_i}{\sigma_i} \right)^2 - 2\left(\frac{t_{i+l} - \mu_i}{\sigma_i} \right)\left(\frac{t_{j+l} - \mu_j}{\sigma_j} \right) + \left( \frac{t_{j+l} - \mu_j}{\sigma_j} \right)^2 \right) \\
&= \sum_{l=0}^{m-1} \left( \frac{t_{i+l}^2  - 2 t_{i+l}\mu_i + (\mu_i)^2}{(\sigma_i)^2} - 2\left(\frac{t_{i+l} t_{j+l}  - \mu_i t_{j+l} - t_{i+l} \mu_j + \mu_j \mu_i}{\sigma_i \sigma_j} \right) + \frac{t_{j+l}^2  - 2 t_{j+l}\mu_j + (\mu_j)^2}{(\sigma_j)^2}\right) \\
\end{align*}
Let
$$
A_i = \sum_{l=0}^{m-1} t_{i+l}, \qquad B_j= \sum_{l=0}^{m-1} t_{j+l},\qquad \mathbf{A}_i =\sum_{l=0}^{m-1} t_{i+l}^2, \qquad \mathbf{B}_j =\sum_{l=0}^{m-1} t_{j+l}^2, \qquad \mathbf{C_{i,j}}= \sum_{l=0}^{m-1} t_{i+l} t_{j+l}.
$$
Then, we have:
$$
\mu_i=\frac{1}{m} A_i, \qquad \mu_j = \frac{1}{m} B_j
$$
$$
(\sigma_i)^2 = \frac{1}{m} \mathbf{A}_i- \frac{1}{m^2} A_i^2, \qquad (\sigma_j)^2= \frac{1}{m} \mathbf{B}_j- \frac{1}{m^2} B^2_j.
$$
Then, the z-normalized Euclidean distance can be written as: 
\begin{align*}
DZ_{i,j}^2
&= \sum_{l=0}^{m-1} \left( \frac{t_{i+l}^2  - 2 t_{i+l}\mu_i + (\mu_i)^2}{(\sigma_i)^2} - 2\left(\frac{t_{i+l} b_{j+l}  - \mu_i t_{j+l} - t_{i+l} \mu_j + \mu_j \mu_i}{\sigma_i \sigma_j} \right) + \frac{t_{j+l}^2  - 2 t_{j+l}\mu_j + (\mu_j)^2}{(\sigma_j)^2}\right) \\
&= \frac{\mathbf{A}_i - 2 A_i^2 \frac{1}{m} + \frac{A_i^2}{m} }{\frac{1}{m} \mathbf{A}_i- \frac{1}{m^2} A_i^2} - 2 \times  \frac{ \mathbf{C_{i,j}} - \frac{2}{m} A_{i} B_j + \frac{A_i B_j}{m}}{\sqrt{(\frac{1}{m} \mathbf{A}_i- \frac{1}{m^2} A_i^2)( \frac{1}{m} \mathbf{B}_j- \frac{1}{m^2} B_j^2)}}
+\frac{\mathbf{B}_j - 2 B_j^2 \frac{1}{m} + \frac{B_j^2}{m} }{\frac{1}{m} \mathbf{B}_j- \frac{1}{m^2} B^2_j}\\
&= 2 m - 2 \times  \frac{ m^2 \mathbf{C_{i,j}} - m A_{i} B_j }{\sqrt{( m \mathbf{A}_i-  A_i^2)(m \mathbf{B}_j-  B_j^2)}}= 2 m\left( 1 -   \frac{  \mathbf{C_{i,j}} - \frac{1}{m} A_{i} B_j }{\sqrt{(  \mathbf{A}_i- \frac{1}{m} A_i^2)( \mathbf{B}_j-  \frac{1}{m} B_j^2)}}\right).
\end{align*}

$\Box$

As mentioned in Subsection~\ref{acamp_optimizations}, by taking
\begin{equation}
\label{eq:F}
F_{i,j} = \frac{ (A_{i} B_j -m \mathbf{C_{i,j}}) \times |   A_{i} B_j -m \mathbf{C_{i,j}}| }{( \mathbf{A}_i-  \frac{1}{m} A_i^2)( \mathbf{B}_j- \frac{1}{m} B_j)},
\end{equation}
we have $DZ_{i,j}=2m + 2 \text{sign}(F_{i,j}) \times \sqrt{|F_{i,j}|}$ and
we can use the following equivalence in our algorithm:
$$
DZ_{i,j} > DZ_{i,k} \Leftrightarrow F_{i,j}> F_{i,k}.
$$

\twocolumn

\section{Appendix B: Pseudo-code of AAMP algorithm for p-norm distance} \label{aamp_pnorm_pseudo}
Algorithm \ref{aamp_algo_pnorm} shows the pseudo-code of AAMP algorithm for computing the matrix profile while using the p-norm distance for creating the matrix profile. 

\begin{algorithm}
\DontPrintSemicolon
\KwIn{T: time series; n:  length of time series; m: subsequence length}
\KwOut{P: Matrix profile; }

\Begin{

\For{i=1 to n}{

    P[i] = $\infty$ ; // initialize the matrix profile
    } 

\For{k=1 to n-m+1}{
    $dist = PNORM\_Distance (T_{1,m}, T_{k, m})$ // compute the distance between $T_{1,m}$, $T_{k, m}$ 
   
   \If{dist < P[1]}{
        P[1] = dist;
    }
    
    \If{dist < P[k]}{
        P[k] = dist;
        
   }
    
    \For{i=2 to n - m + 1 - k}{
    
        $dist = \sqrt[p]{((dist)^p -  (t_{i-1} - t_{i - 1 + k})^p + (t_{i+m-1} - t_{i + m + k - 1} )^p }$ 
        
        \If{dist < P[i]}{
            P[i] = dist;
            
        }
        
        \If{dist < P[i+k]}{
            P[i+k] = dist;
            
        }
    } 
    }
}

\caption{AAMP algorithm for p-norm distance \label{aamp_algo_pnorm}}	
\end{algorithm}

\end{document}